# 3D Human Pose Estimation for Free-from and Moving Activities Using WiFi

**YILI REN,** Florida State University, USA
**ZI WANG,** Florida State University, USA
**YICHAO WANG,** Florida State University, USA
**SHENG TAN,** Trinity University, USA
**YINGYING CHEN,** Rutgers University, USA
**JIE YANG,** Florida State University, USA

This paper presents GoPose, a 3D skeleton-based human pose estimation system that uses WiFi devices at home. Our system leverages the WiFi signals reflected off the human body for 3D pose estimation. In contrast to prior systems that need specialized hardware or dedicated sensors, our system does not require a user to wear or carry any sensors and can reuse the WiFi devices that already exist in a home environment for mass adoption. To realize such a system, we leverage the 2D AoA spectrum of the signals reflected from the human body and the deep learning techniques. In particular, the 2D AoA spectrum is proposed to locate different parts of the human body as well as to enable environment-independent pose estimation. Deep learning is incorporated to model the complex relationship between the 2D AoA spectrums and the 3D skeletons of the human body for pose tracking. Our evaluation results show GoPose achieves around 4.7cm of accuracy under various scenarios including tracking unseen activities and under NLoS scenarios.

## 1 INTRODUCTION

Estimating the human pose is gaining increasing attention as the human body offers a high degree of freedom for human-computer interactions (HCI). It is a crucial building block to support a variety of emerging applications in smart home, such as virtual/augmented reality [7, 36], interactive exergaming [43, 49], well-being [24, 64], and exercise monitoring [44, 46]. Traditional human pose estimation systems mainly rely on either computer vision technique that requires the installation of specialized cameras (e.g., RGB or infrared cameras) [10, 61], or wearable approach where users wear or carry dedicated sensors (e.g., IMU sensors, RFID) [19, 50]. However, the vision-based systems cannot work in non-line-of-sight (NLoS) and poor lighting scenarios, for example when the

Authors' addresses: Yili Ren, Florida State University, 1017 Academic Way, Tallahassee, FL, 32306, USA, ren@cs.fsu.edu; Zi Wang, Florida State University, 1017 Academic Way, Tallahassee, FL, 32306, USA, ziwang@cs.fsu.edu; Yichao Wang, Florida State University, 1017 Academic Way, Tallahassee, FL, 32306, USA, yichaowang@cs.fsu.edu; Sheng Tan, Trinity University, One Trinity Place, San Antonio, TX, 78212, USA, stan@trinity.edu; Yingying Chen, Rutgers University, 671 Route 1, North Brunswick, NJ, 08902, USA, yingche@scarletmail.rutgers.edu; Jie Yang, Florida State University, 1017 Academic Way, Tallahassee, FL, 32306, USA, jie.yang@cs.fsu.edu.

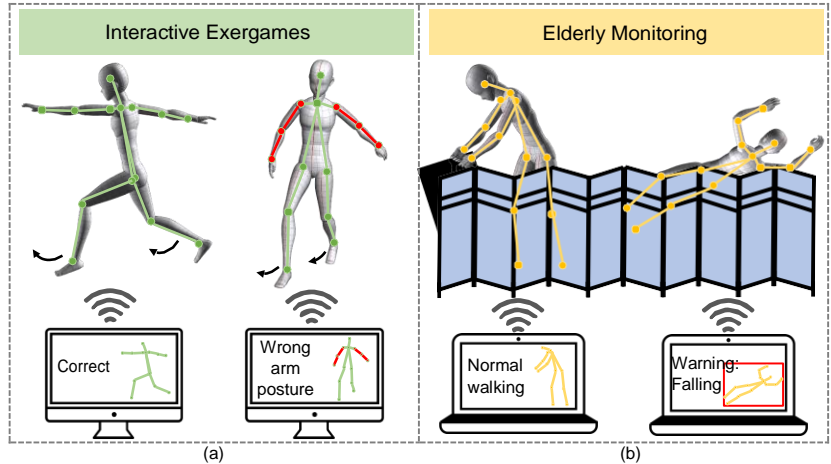

Fig. 1. Motivating examples of GoPose. (a) The computer tracks the 3D poses of the moving user in interactive exergames and tells the user whether her/his postures are correct. (b) The computer provides elderly monitoring by estimating the elders' 3D poses even they walk behind a screen or wall (i.e., NLoS) and sends alerts when a falling or injury occurs.

person is behind a folding screen or in a dark environment, whereas the wearable systems could be inconvenient as they require explicit user involvement. Moreover, the necessity for dedicated devices plus the cost of the hardware in these systems dampens the likelihood of mass adoption.

More recently, Radio Frequency (RF) based sensing becomes an appealing alternative for human pose estimation. It analyzes the RF signals reflected off the human body for activity and human pose tracking, and thus does not require a user to wear or carry any sensors. It also works under NLoS scenarios as the RF signals can penetrate folding screens and walls when compared to the vision-based approach. Existing work in RF-based human sensing uses either specialized hardware (e.g., USRP, FMCW RADAR) [31, 63] or commodity WiFi devices [25, 42, 53]. As the specialized hardware-based systems are less attractive to consumer-oriented use due to their high hardware cost, we focus on the commodity WiFi-based approach as it could be enabled with a simple software update by reusing existing WiFi devices in smart home environments [41].

In this work, we propose GoPose, a 3D skeleton-based human pose estimation system by reusing WiFi devices in a home environment. Unlike the prior WiFi-based 3D human pose estimations that only work for a set of predefined activities [13] performed at a fixed position [37], our system works for unseen activities even when the user is moving around, offering on-the-go pose tracking for unseen activities. It is because our system can extract the two-dimensional (2D) angle of arrival (AoA) of the incident signals, which can represent the spatial information of different body parts or joints regardless of activities or user positions. A deep learning model is then used to identify the joints and transfer the 2D AoA spatial information to the joint locations in physical space. WiPose extracts the features which cannot directly represent the spatial information of the body joints. It utilizes the deep learning model to map the features that are less related to the spatial information to the joint locations. Such a mapping relationship could be weak when the user is moving around or performing unseen activities. As shown in Figure 1, GoPose could be utilized to digitize a user's 3D full-body motions into a set of body joints to enable new interactive experiences beyond the traditional computer vision and touch-based human-computer input. It works when people are mobile and across occlusions, where the light of sight is blocked by a folding screen or walls. As GoPose could reuse WiFi devices, it does not incur an additional cost, and thus is promising for mass adoption for end-users in smart homes.

Estimating 3D human pose solely from the WiFi signals bounced off the human body faces unique challenges. First, unlike the USRP or FMCW RADAR that offers accurate spatial information (e.g., the location and shape of objects) [62, 63], the Channel State Information (CSI) data directly exported from the off-the-shelf WiFi devices does not provide any spatial information of the human body. To tackle this challenge, we leverage the two-dimensional angle of arrival of the incident signals derived from the non-linearly spaced antennas to provide spatial information for pinpointing the human body. Moreover, we propose to combine both the spatial diversity at the transmitter and the frequency diversity of WiFi OFDM subcarriers to increase the spatial resolution of 2D AoA for differentiating signals reflected from different parts of the human body.

Another challenge is that as the received WiFi signal is dominated by the signal reflected from the indoor environments, such as those from the walls and furniture, how to make the human pose estimation system independent of the environments it operates in? That is once the system is configured in one environment, it should work well across different environments, for example after move to a different house. To handle this challenge, we leverage the spatial characteristic of the 2D AoA spectrum to separate the signals bounced off the human body from the ones reflected from the static environments. In particular, we subtract the 2D AoA spectrum of the static environment from the ones we extracted when one or more users are performing activities. In addition, we propose to combine the 2D AoA spectrum of multiple packets at multiple receivers to resolve the issue of specularity of the human body, i.e., one received WiFi packet only captures a subset of body motions at a particular direction.

The next challenge is to model the complex relationship between the 2D AoA spectrums and the 3D skeletons of the human body. The complexity is very high as the human body has a high degree of freedom and the user could be moving around at different locations with different orientations. Instead of using analytic kinematic models, we leverage the deep learning models of the convolutional neural network (CNN) and the Long Short-Term Memory (LSTM) to abstract the 3D human pose from the 2D AoA spectrums. In particular, the CNN is a useful technique to extract spatial dynamics (e.g., the locations of multiple limbs and the torso), whereas the LSTM is a special kind of recurrent neural network (RNN) that models and estimates temporal dynamics of human poses (e.g., trajectories of limbs and torso).

We experimentally evaluate the GoPose in different home environments with various activities performed by different users. We conduct experiments under non-line-of-sight scenarios, different distances between WiFi devices, and multi-person scenarios. Results show that our system is highly accurate in constructing 3D human poses for moving users even for unseen activities. The main contributions of our work can be summarized as follows:

- We propose a 3D human pose estimation system for moving user and unseen activities by leveraging WiFi devices. The proposed system does not require any dedicated or specialized sensors and can work under NLoS scenarios.
- We estimate the human pose based on the 2D AoA spectrums derived from the non-linearly spaced antennas. The 2D AoA spectrum offers unique advantages including providing spatial information of the human body and enabling environment-independent human pose estimation.
- We leverage deep learning (i.e., CNN and LSTM) to model the 2D AoA spectrums and the human body for inferring 3D skeletons of human pose. Experimental results show that GoPose achieves around 4.5 cm of accuracy under various scenarios including tracking unseen activities and under the NLoS scenarios.

## 2 RELATED WORK

Existing work of human pose estimation can be divided into three categories: computer vision-based, wearable sensor-based, and RF signal-based.

**Computer vision-based.** There exist many systems for 2D or 3D body skeleton and pose estimation by utilizing images or videos. For example, recent conventional RGB camera-based 2D body pose estimation systems [4, 10, 29] have made great progress by leveraging the deep learning models and annotated pose databases. Meanwhile, 3D pose estimation has also attracted growing interest from researchers. For instance, Sigal *et al.* [40] presented a baseline algorithm for 3D human pose estimation by leveraging an optimized relatively standard Bayesian framework, whereas Pavllo *et al.* [30] proposed a fully convolutional model based on semi-supervised training video data over two dimensions key points. VNect [26] achieves full global 3D human pose tracking with only one RGB camera, while Kanazawa *et al.* [14] reconstruct a full 3D mesh human body model with a single RGB image. Moreover, LiSense [21] utilizes visible light communication to achieves concurrent 3D human skeleton reconstruction and real-time data communication, whereas Ahuja *et al.* [1] proposed to leverage multiple cameras and sensors on the smartphone to approximate the full-body pose while on the go. Similarly, Simo [2] uses cameras of the smartphone to track body movements, whereas cameras were installed below the floor to track human poses in one system [3]. Additionally, the commercial depth/infrared cameras are proposed to track 3D human pose, for example, in Microsoft Kinect [27] and Leap Motion [11, 48]). These computer vision-based approaches, however, cannot work in non-line-of-sight (NLoS) and poor lighting scenarios. In addition, the vision-based approaches often involve user privacy concerns and incur a non-negligible cost.

**Wearable sensor-based.** The wearable sensor-based approaches require the user to carry or wear one or multiple dedicated sensors. For example, existing systems [6, 15] can track the movement of arms by attaching various sensors on the upper limbs of the user. Moreover, 3D hand pose could be estimated by using a wrist-worn camera [55] or attachable electromagnets [5], whereas the full-body motions could be reconstructed by attaching the IMU sensors to various positions of the body [12, 47]. In addition, there are several systems that utilize a single wearable sensor for body motion tracking [33, 34]. For instance, RecoFit [28] can provide real-time feedback and post-workout analysis for strength-training exercises by leveraging a single sensor on the arm, while ArmTrack [39] utilized a single smartwatch to rebuild the arm motions. However, these wearable sensor-based systems all rely on dedicated sensors that need to be attached to or worn by the user, which could be inconvenient and cumbersome under daily life scenarios and incur non-negligible installation overhead.

**RF signal-based.** Many research efforts have been dedicated to human pose tracking leveraging RF signals [41]. For instance, Zhao *et al.* proposed RF-Pose [62] that leverages a teacher-student network to estimate 2D human pose through the walls. They further proposed RF-Pose3D [63], which could infer 3D human body skeletons based on a convolutional neural network model. However, these systems rely on the specialized hardware that emits FMCW signals across a large bandwidth, which is dozens of times wider than that of the WiFi bandwidth. Moreover, it requires a carefully designed and synchronized T-shape antenna array to obtain accurate spatial information, which contains four antennas and sixteen antennas for vertical array and horizontal array, respectively. Moreover, Zhang *et al.* presented Wall++ [60], a sensing approach that can estimate the body pose of users but requires patterning large electrodes onto a wall using conductive paint. However, these systems leveraging specialized hardware are less scalable for mass adoption due to the cost of hardware and the overhead of installation.

To enable potential mass adoption, many systems utilize WiFi devices that can be found in a home environment to enable a wide range of applications including large-scale activity recognition [32, 53] (e.g., daily activities, dancing movements), indoor locations [23, 56, 57, 65], small-scale motion sensing [25, 42] (e.g., vital sign, finger gesture), and object sensing [35, 45] (e.g., fruit ripeness, liquid level). For example, E-eyes [53] and WiFinger [42] are among the first work to leverage commodity WiFi to classify different daily activities and finger gestures respectively, whereas Liu *et al.* [25] and Tan *et al.* [45] are among the first to perform vital signs and object sensing (i.e., FruitSense [45]), respectively. For human pose estimation, Person-in-WiFi [51] rebuilds the 2D skeleton for pose estimation. The system heavily relies on the features such as Part Affinity Fields [4] and Segmentation Masks [10], which are only suitable for 2D scenarios. WiPose [13] is a more recent work using commercial off-the-shelf

WiFi devices to track the 3D human pose. It however only works well for a set of predefined activities performed. Winect [37] is the most recent system that works for free-form human activities tracking but is limited to the activities performed at a fixed location. In our work, we propose a 3D skeleton-based human pose estimation system that offers on-the-go pose tracking even for unseen activities.

**2D AoA.** It is worth noting that there are many research efforts related to 2D AoA. For example, Lee *et al.* proposed a low-complexity estimation algorithm [18] for the estimation of 2D AoA using a pair of uniform circular arrays. In addition, Li *et al.* proposed a 2D AoA estimation model [20] based on the motion of a 1D nested array. Wei *et al.* presented a method [54] for pair-matching of elevation and azimuth angles in 2D AoA estimation with the L-shaped array. Wang *et al.* proposed a framework [52] that occupies a larger and provided better estimation of 2D AoA performance for multiple-input multiple-output (MIMO) radar. mmEye [59] is a super-resolution imaging system toward a mmWave on commodity 60GHz WiFi devices. It developed a super-resolution imaging algorithm based on Multiple Signal Classification (MUSIC) and 2D AoA. However, this system achieves 2D human imaging instead of 3D human pose estimation.

To the best of our knowledge, none of them use 2D AoA to estimate the 3D human pose with WiFi signals. In our application, we further leverage the spatial diversity at transmitting antennas and the frequency diversity of OFDM subcarriers of WiFi to improve the resolution of the 2D AoA. We note that some of these 2D AoA techniques could be incorporated into our system.

## 3 PRELIMINARIES

In this section, we present the preliminary of WiFi sensing, and discuss 1D and 2D AoA estimation of the incident signals and their limitations.

### 3.1 WiFi Sensing

WiFi has been evolving from providing laptop connectivity to connecting smart home devices such as tablets, smartphones, smart speakers, refrigerators, and smart TVs to home networks and the Internet. It has resulted in a large number of WiFi devices, which provides the opportunity to extend WiFi's capabilities beyond communication, particularly in sensing the physical environment. As the WiFi signals travel through space, they interact with the human body, and any human activities, either small scale or large scale, affect the signal propagation. With measurable changes in the received WiFi signals, human activities in the physical environment thus could be inferred.

Moreover, with the advanced WiFi technology, WiFi radio offers channel state information (CSI) (i.e., the sampled version of the channel frequency response) to estimate the channel condition for fast and reliable communication. The CSI thus could be directly exported from the network interface card (NIC) to measure the changes in the CSI amplitude and phase for inferring human activities. In particular, current 802.11a/g/n/ac employs OFDM technology, which partitions the relatively wideband WiFi channel into 52 subcarriers and provides detailed CSI for each subcarrier.

The CSI directly exported from the commodity WiFi devices only provides information on how the wireless channel was interrupted by human activities. It however offers no spatial information regarding human activities, such as the location and the shape of the human body. We thus turn to exam the AoA of the incident signals at the antenna array of WiFi devices to derive spatial information for human pose estimation.

### 3.2 1D AoA Estimation

The one-dimensional (1D) angle of arrival (AoA) of the incident signals (e.g., the LoS signals and the reflected signals from the human and environment) could be derived if the WiFi receiver is equipped with a linear antenna

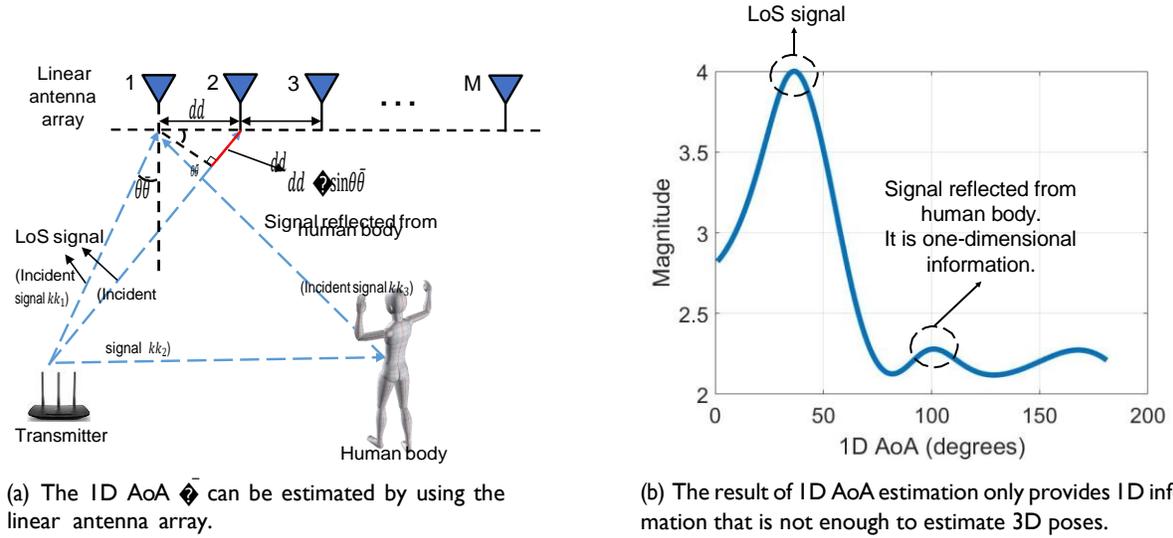

(a) The 1D AoA $\bar{\theta}$ can be estimated by using the linear antenna array.

(b) The result of 1D AoA estimation only provides 1D information that is not enough to estimate 3D poses.

Fig. 2. Linear antenna array for 1D AoA estimation.

array. This is because the incident signals incur different phase shifts across the antennas in an array, as shown in Figure 2(a). Let's assume that there are $M$ antennas in a uniform linear array, and the distance between adjacent antennas is $d$ ($d$ is a half wavelength). Taking the LoS signals in Figure 2(a) as an example, the LoS signal $k_1$ propagates to the antenna array with an AoA of $\bar{\theta}$, and the path difference of the LoS signal between adjacent antennas (e.g., the difference between $k_1$'s path and $k_2$'s path) is $d \cdot \sin(\bar{\theta})$. Then, relative to the first antenna in the array, the phase shift introduced at the $m_{th}$ antenna is

$$\Phi(\bar{\theta})_m = 2\pi f(m-1) \cdot d \cdot \sin(\bar{\theta})/c, \qquad (1)$$

where $f$ is the frequency of the signal and $c$ is the speed of light. We thus can denote the introduced complex exponential phase shift as a function of the 1D AoA: $\Phi(\bar{\theta}) = e^{-j2\pi f d \sin(\bar{\theta})/c}$. For the LoS signal, the phase shifts across the $M$ antennas in the uniform linear array can be denote as

$$\mathbf{a}(\bar{\theta}) = [1 \; \Phi(\bar{\theta}) \; ... \; \Phi(\bar{\theta})^{M-1}]^T, \qquad (2)$$

where $\mathbf{a}$ is called as the steering vector.

The indoor multipath environment includes numerous incident signals from different directions, for example, the LoS signal (e.g., $k_1$ or $k_2$) and signals reflected from the human body (e.g., $k_3$) in Figure 2(a). Therefore, when $K$ incident signals from different angles arriving at the uniform linear antenna array of $M$ antennas, the received signal at each antenna is the superposition of all incident signals, and the corresponding steering vectors form the steering matrix $[\mathbf{a}(\bar{\theta}_1), \ldots, \mathbf{a}(\bar{\theta}_K)]$. Based on this, AoAs of these incident signals can be derived by using the MUSIC algorithm [38].

Figure 2(b) shows one example of the estimated 1D AoA of the LoS signal and the signal reflected from the human body. We can observe that 1D AoA offers the direction of the indicant signal with respect to the antenna array. For example, the peak at around 40 degrees indicates the LoS signal, whereas the second peak at around 100 degrees represents the signal reflected off the human body. However, 1D AoA only provides very limited

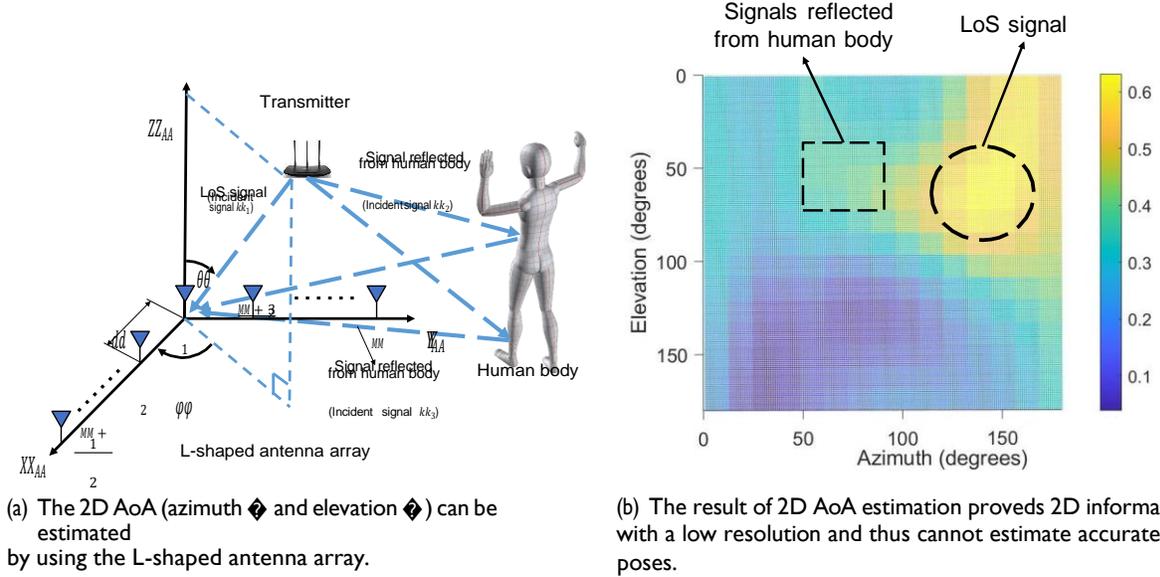

(a) The 2D AoA (azimuth φ and elevation θ) can be estimated by using the L-shaped antenna array.

(b) The result of 2D AoA estimation proveds 2D information with a low resolution and thus cannot estimate accurate 3D poses.

Fig. 3. L-shaped antenna array for 2D AoA estimation.

spatial information, which is insufficient for tracking multiple human limbs. For instance, it cannot provide the position of a person in a 2D space, not to mention to pinpoint the spatial locations of multiple human limbs.

### 3.3 2D AoA Estimation

In our work, we extend the 1D AoA estimation to 2D AoA estimation with the help of the non-linearly spaced antennas at the WiFi device. This provides the opportunity to distinguish different signals in the 2D space instead of the 1D space. In particular, we leverage an L-shaped antenna array at the receiver to derive both the azimuth angle $\varphi$ and the elevation angle $\theta$ of the incident signals [9]. As shown in Figure 3(a), we assume the L-shaped antenna array has $M$ antennas separated by distance $d$ at the receiver, and these antennas form the $X_A$-$Y_A$-$Z_A$ coordinate system. Then, we can formulate the 2D AoA estimation as follows.

As shown in Figure 3(a), we assume that there are $K$ incident signals including the LoS signal (e.g., $k_1$), and signals reflected from the human body (e.g., $k_2$ and $k_3$). These signals are arriving at the L-shaped antenna array in the $X_A$-$Y_A$-$Z_A$ coordinates. Similar to Equation 1, the phase shift for the $k^{th}$ incident signal on the $m^{th}$ antenna can be rewritten as

$$\Phi_m(\varphi_k, \theta_k) = e^{-j2\pi f \sin(\theta_k)[X_{Am}\cos(\varphi_k)+Y_{Am}\sin(\varphi_k)]/c}, \quad (3)$$

where $(X_{Am}, Y_{Am})$ denote the $m^{th}$ antenna's coordinates. For example, the first antenna is located at $(0,0)$ and the second antenna is located at $(d, 0)$. Correspondingly, the steering vector for 2D AoA can be denoted as

$$\mathbf{a}(\varphi, \theta) = [\Phi_1(\varphi_k, \theta_k)\ \Phi_2(\varphi_k, \theta_k)\ ...\ \Phi_M(\varphi_k, \theta_k)]^T, \quad (4)$$

which is similar to Equation 2. Thus, the steering matrix for 2D AoA is $[\mathbf{a}(\varphi_1, \theta_1), \ldots, \mathbf{a}(\varphi_K, \theta_K)]$. Likewise, we can leverage the MUSIC algorithm to calculate the 2D AoA spatial spectrum.

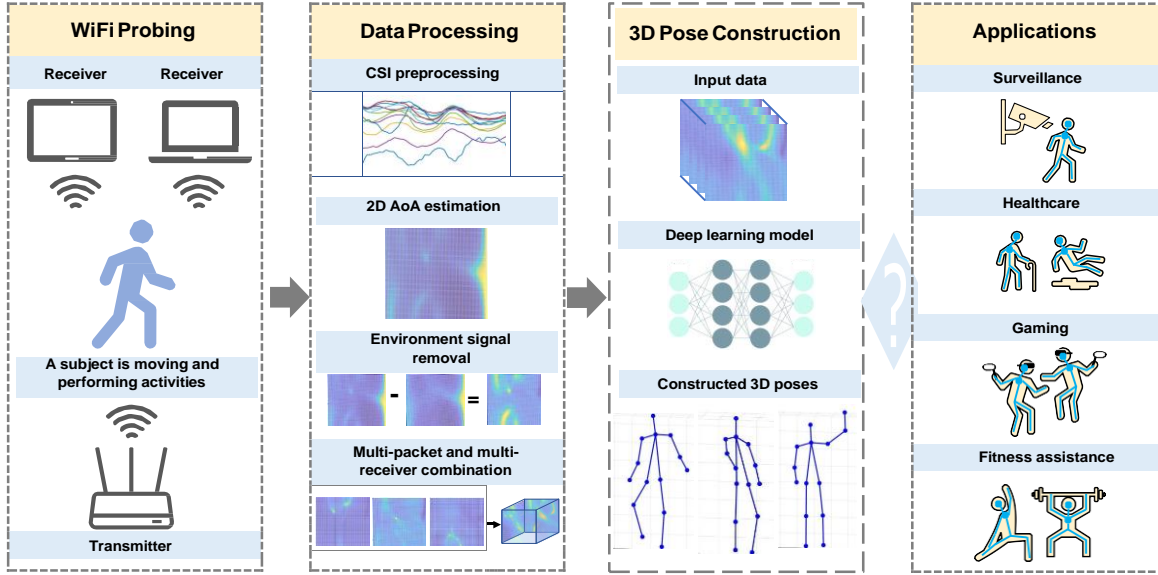

Fig. 4. System overview.

Figure 3(b) shows the result of the 2D AoA estimation with commodity WiFi devices. We can observe that the resolution is low as the commodity WiFi device only support up to 3 antennas. Specifically, there are two antennas on each of the $X_A$-axis and $Y_A$-axis with one antenna shared by two axes. We can observe from Figure 3(b), the fuzzy peak with strong spectrum energy represents the 2D location of the transmitter, whereas the area with lower energy indicates the 2D location of the signals reflected from the human body. Although the 2D AoA derived from commodity WiFi devices can provide the approximate location of the human body in 2D space, it is not able to differentiate signals reflected from different parts of the human body, such as these from the torso (i.e., signal $k_2$) or from the leg (i.e., signal $k_3$). This is because the hardware constraints of the commodity WiFi result in a very low resolution of the 2D AoA spectrum. To overcome such a limitation, we further propose to improve the resolution of the 2D AoA spectrum by combining both the spatial diversity at the transmitter and the frequency diversity of WiFi OFDM subcarriers. We present the details in Section 4.2.

## SYSTEM DESIGN

In this section, we discuss the system design, present the methods to improve the resolution of the 2D AoA spectrum and remove the environment effects, and describe the design of the deep learning models.

### 4.1 System Overview

The basic idea of our system is to leverage the spatial information of the 2D AoA spectrum and deep learning to model the complex 3D skeletons of the human body for 3D pose estimation. As illustrated in Figure 4, a WiFi transmitter sends out signals to multiple WiFi receivers to probe human activities. The system takes as input time-series CSI measurements, which can be exported from the NICs of the commodity WiFi devices. The CSI measurements are exported for 30 subcarriers on each WiFi link. The system can benefit from the CSI measurements from existing traffic across these links, or the system can also generate periodic traffic for sensing purposes. This data is then preprocessed to remove noises by using a linear fit method proposed in the literature.

The core of our system, GoPose, is 2D AoA extraction and the 3D pose construction. The 2D AoA extraction encompasses three different components to address the issues of spatial resolution, environment independence, and the specularity of the human body. The system first combines both the spatial diversity and the frequency diversity to increase the resolution of 2D AoA for differentiating signals reflected from different parts of the human body. It then goes through static environment removal to filter out the signals reflected from the indoor environments. After that, the system combines the 2D AoA spectrum of multiple packets at multiple receivers to resolve the issue of specularity of the human body (i.e., one packet can only capture a subset of motions of the human body).

Next, our system leverages the deep learning models of CNN and LSTM to construct the 3D pose of the human body based on the 2D AoA spectrum. CNN is used to capture the spatial feature of the human body parts, while the LSTM is utilized to estimate the temporal feature of the motions. GoPose offers on-the-go pose tracking for unseen activities. As it relies on the WiFi signal reflections for human pose tracking, it does not require the user to wear or carry any devices and works under NLoS scenarios. It could also be enabled with a simple software update for mass adoption as it can reuse existing WiFi devices in a home environment.

## 4.2 Improving 2D AoA Estimation

As the limited number of antennas on commodity WiFi receivers (e.g., Intel 5300 card has up to only three antennas) provides insufficient 2D AoA resolution for 3D human pose estimation, we seek helps from both the spatial diversity at the transmitter and the frequency diversity of the WiFi OFDM subcarriers. In particular, existing approaches [17, 22] only utilize the frequency diversity of OFDM subcarriers to calculate the time of flight (ToF) to improve AoA estimation. In our work, we improve the 2D AoA estimation by leveraging both the spatial diversity in three transmitting antennas as well as the frequency diversity of thirty OFDM subcarriers. The spatial diversity in three transmitting antennas can introduce phase shifts due to the angle of departure (AoD), while the frequency diversity of OFDM subcarriers can result in phase shifts with respect to time of flight (ToF). Thus, we can jointly estimate 2D AoA, AoD, and ToF by leveraging both the spatial and frequency diversities to dramatically improve the resolution of the 2D AoA spectrum.

Specifically, we utilize CSI measurements of the WiFi signals across all OFDM subcarriers transmitted from multiple transmitting antennas and received at multiple receiving antennas to generate a large number of virtual sensing elements. In our implementation, for each subarray (i.e., on $X\_A$-axis or $Y\_A$-axis in the $X\_A$-$Y\_A$-$Z\_A$ coordinates), we have two receiving antennas, three transmitting antennas, and thirty subcarriers. Thus, there are in total 180 sensing elements for each axis. It provides three times better resolution than the ones estimated in the existing work [17, 22]. Or it offers ninety times better resolution when compared to 1D AoA estimation. The spatial and frequency diversities thus result in sufficient information that allows our system to jointly estimate high-resolution 2D AoA (azimuth and elevation), AoD, and ToF simultaneously. The information can be combined together to improve the resolution of 2D AoA estimation.

Figure 5 illustrates that we can separate multipath signals in 2D space and capture different parts of the human body of the moving subject with the improved 2D AoA spectrum. Figure 5($a_1$) shows multipath signals including the LoS signal, the signal reflected by the wall, and the signals reflected from different parts of the human body. The resulted 2D AoA spectrum is shown in Figure 5($a_2$). We can observe that the improved 2D AoA spectrum can be used to differentiate the multipath signals, such as the signals that come from the LoS, environment, and human body reflections. We can also observe that the signals reflected from different parts of the human body (e.g., arms, legs, and torso) are located at different spatial locations, as shown in Figure 5($c_2$).

Specifically, we formulate the improved 2D AoA estimation as follows. We assume the signal emitted from a linear transmitting antenna array will be received with a phase shift $\Gamma(\omega)$, which is the function of AoD. For the $k^{th}$ path with AoD $\omega_k$, the phase difference across transmitting antennas is given by:

$$\Gamma(\omega_k) = e^{-j2\pi f d' \sin(\omega_k)/c}, \quad (5)$$

where $d'$ is the distance between transmitting antennas.

Current IEEE 802.11 standard adopts OFDM technology, wherein the data is transmitted over multiple subcarriers. Thus, for equally spaced OFDM subcarriers, the $k^{th}$ path with ToF of $\tau_k$ introduces a phase shift across two consecutive OFDM subcarriers with $f_\delta$ frequency difference that can be represented as follows:

$$\Omega(\tau_k) = e^{-j2\pi f_\delta \tau_k/c}. \quad (6)$$

We then jointly estimate the 2D AoA (azimuth and elevation), AoD, and ToF by defining the sensing element array from all the subcarriers of all the receiving and transmitting antenna pairs. The overall attenuation and phase shift introduced by the channel measured at each subcarrier by each antenna is reported as the CSI in a $R \times S \times V$ format ($R$ represents the number of receiving antennas, $S$ represents the number of transmitting antennas, and $V$ represents the number of subcarriers). Therefore, the sensor array can be constructed through stacking CSI from all the subcarriers, resulting in a total number of $R \times S \times V$ sensors. Compared to Equation 4, the new steering vector $\mathbf{a}(\varphi, \theta, \omega, \tau)$ is formed by phase difference introduced at each of the sensors and is given by:

$$\mathbf{a'}(\varphi, \theta, \tau) = [1, \ldots, \Omega_\tau^{V-1}, \Phi_{(\varphi,\theta)}, \ldots, \Omega_\tau^{V-1}\Phi_{(\varphi,\theta)}, \ldots, \Phi_{(\varphi,\theta)}^{R-1}, \ldots, \Omega_\tau^{V-1}\Phi_{(\varphi,\theta)}^{R-1}]^T, \quad (7)$$

$$\mathbf{a}(\varphi, \theta, \omega, \tau) = [\mathbf{a'}_{(\varphi,\theta,\tau)}, \Gamma_\omega \mathbf{a'}_{(\varphi,\theta,\tau)}, \ldots, \Gamma_\omega^{S-1}\mathbf{a'}_{(\varphi,\theta,\tau)}]^T, \quad (8)$$

where $\Gamma_\omega$, $\Omega_\tau$, $\Phi_{(\varphi,\theta)}$ and $\mathbf{a'}_{(\varphi,\theta,\tau)}$ are the abbreviations of $\Gamma(\omega)$, $\Omega(\tau)$, $\Phi(\varphi, \theta)$, and $\mathbf{a'}(\varphi, \theta, \tau)$ respectively. Therefore, the received signal can be constructed using the above steering vector. Parameters of azimuth, elevation, AoD, and ToF that maximize the spatial spectrum function [38] can be estimated by:

$$P(\varphi, \theta, \omega, \tau)_{Improve} = \frac{1}{\mathbf{a}^H(\varphi, \theta, \omega, \tau)\mathbf{E}_N\mathbf{E}_N^H\mathbf{a}(\varphi, \theta, \omega, \tau)}. \quad (9)$$

Lastly, we derive the azimuth-elevation 2D AoA power spectrum by accumulating the AoA values in dimensions of ToF and AoD.

### 4.3 Static Environment Removal

As the 2D AoA spectrum provides spatial information of the multipath signals, we can leverage such information to remove the LoS signal and the signals reflected from the static environment for environment-independent 3D pose estimation. In particular, we propose to subtract the 2D AoA spectrum of the static environment from the ones we extracted with human activities. Then, the 2D AoA spectrum mainly reflects the signals bounced off the human body and thus is independent of the signals reflected from the static environment. More specifically, we first calculate the 2D AoA spectrum of the static environment from multiple CSI packets. For example, Figure 5($b_1$) shows the signals of the static environment, which include the LoS signal and the signal reflection from static objects (e.g., wall). The corresponding spectrum is shown in Figure 5($b_2$), in which we can distinguish the 2D locations of the LoS signal and signals reflected from the wall. Note that the spectrum of the static environment should be periodically updated after detecting significant changes in the environment (e.g., the furnishings have been significantly altered) [58].

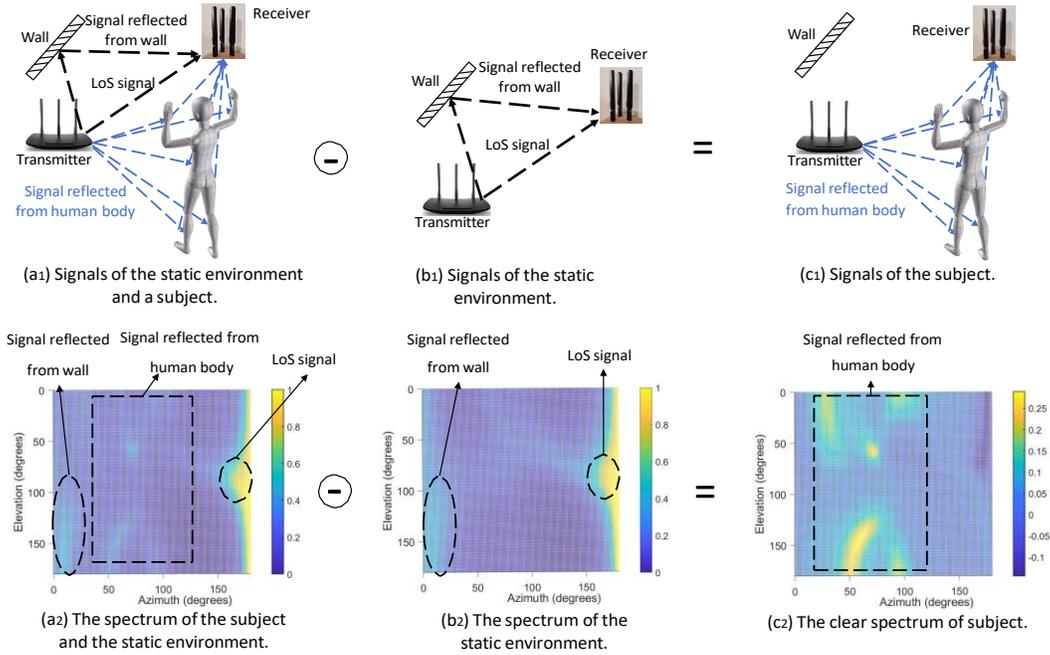

Fig. 5. Static environment removal enables environment-independent sensing.

The 2D AoA spectrum under human activities will also be generated. As shown in Figure 5($a_1$), in which the person is walking towards the receiver while waving his hands. We can observe from the corresponding 2D AoA spectrum, i.e., Figure 5($a_2$), although the signal reflected from the human body is weaker compared to the LoS signal, we still see the signals reflected from different parts of the human body. Next, we subtract the static spectrum from the spectrum under human activities to obtain the 2D AoA spectrum that only reflects the signals bounced off the human body. As shown in Figure 5($c_2$), we can clearly observe the signals reflected from human's different limbs and torso, which is irrelevant to the environments.

It is worth noting that the signals reflected from the human body may bounce off the walls again, resulting in secondary reflection to the receivers. For example, when a person shows up, there might be a signal propagates from the transmitter to the person, then reflected from the person to the wall and eventually received by the receiver after wall reflection. Although such a signal can not be removed from spectrum subtraction, it has little effect on human pose estimation as such it is too weak after the second reflection.

### 4.4 Combining Multiple Packets and Multiple Receivers

The human body is specular with respect to WiFi signals, which means the human body acts as a reflector (i.e., a mirror) instead of a scatterer [62]. This is because the wavelength of the WiFi signal is much larger than the roughness of the surface of the human body. On contrary, the human body acts as a scatterer with respect to visible light as its wavelength is much smaller than the roughness of the surface of the human body. Depending on the orientation of the human body, some WiFi signals may be reflected towards the receiver, while some may be reflected away from the receiver. As a result, the 2D AoA spectrum derived from a single WiFi packet can only capture a small subset of body motions and may miss the majority part of the motions.

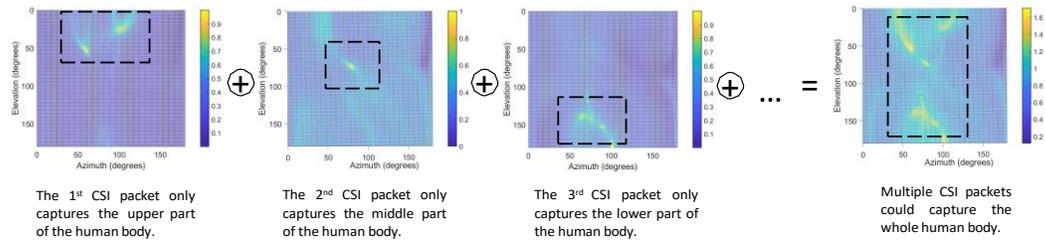

Fig. 6. An example of combining multiple packets to capture the whole human body.

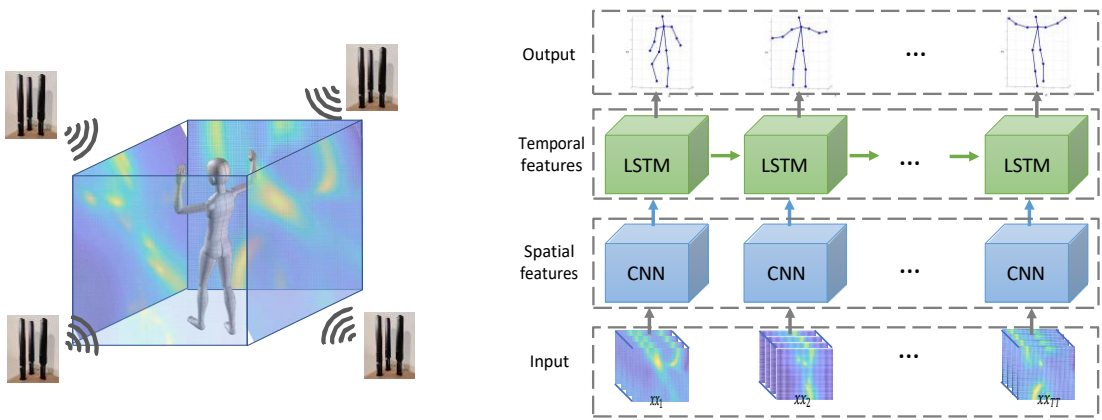

Fig. 7. Using multiple receivers to capture 3D information.

Fig. 8. Deep learning framework.

To resolve this issue, we combine multiple 2D AoA spectrums derived from multiple CSI packets to capture the motions of different parts of the human body. As shown in Figure 6, we can observe that the first spectrum only captures the upper part of the body (e.g., two arms) and other parts are missing. The second spectrum only captures the middle part of the body (e.g., the torso), whereas the third one only has information about the legs. This actually inspired us to take advantage of multiple WiFi packets to describe the full-body movements, as shown in the last spectrum in Figure 6. It is worth noting that the last spectrum is an intuitive example that is superimposed by the previous three spectrums. In our system, we let the deep learning networks learn such information from multiple spectrums derived from multiple packets. In particular, we take a sequence of packets (i.e., 100 packets) as input to estimate one human pose.

To obtain the 3D information from the 2D AoA spectrum, we leverage multiple receivers locations at different positions, as shown in Figure 7. Each receiver is equipped with an L-shaped antenna array and can be used to extract one 2D AoA spectrum. By combing the 2D spectrums from multiple receivers, we are able to recover 3D information of the human pose. As the complexity of the 3D human pose is very high, we leverage the deep learning models instead of analytic kinematic models to infer the 3D human pose based on the 2D AoA spectrums.

## 4.5 Deep Learning Framework

The deep learning framework of GoPose is illustrated in Figure 8. After 2D AoA estimation, we obtain the spectrum with the dimensions of 180 × 180 as we set the range of azimuth and elevation as [0, 180] degrees with a resolution of one degree. Our system also utilizes multiple receivers (e.g., 4 receivers) to capture the motions of the user from different angles. We concatenate the spectrum from four receivers, and derive the tensor with the dimensions of 180×180×4. Moreover, we need to combine multiple spectrums to capture the full-body movements. We thus concatenate 100 packets of each receiver to form a 180 × 180 × 400 matrix, and we denote such a matrix as $x\_t$, which means the input data at time $t$. Then, the whole sequence of input data can be denoted as $[x_1, x_2, ..., x\_t]$. Each input data illustrates the spectrum distribution of one snapshot of the moving user in the physical space, and the continuous input data stream describes how the spectrum varies corresponding to human activities. To fully understand the input data stream, we extract both spatial features from each input data $x\_t$ and the temporal dependencies between $x\_i$ and $x\_j$.

In particular, we first adopt CNN in our deep learning model to extract spatial features from $x\_t$. In our system, the spatial features could be the positions of different parts of the body in the 2D AoA spectrum. More specifically, we utilize stacked six-layer CNNs and use 3D filters for each CNN layer. After the CNN layer, we adopt a batch normalization layer to standardize the inputs to a network and speed up the training. Then, we add a rectified linear unit (ReLU) to add non-linearity. After that, max pooling is applied to down-sample the features. Also, a dropout layer is added to prevent overfitting.

Besides the spatial features, the input data also contains temporal features as the 3D skeleton is dynamic and the movement is consecutive. Recurrent neural networks (RNN) are useful as they can model complex temporal dynamics of sequences. Compared with original RNNs, Long Short-Term Memory (LSTM) is more capable of learning long-term dependencies. After the CNNs, we obtain a sequence of the feature vector, which is then fed into the LSTM. In particular, we utilize a two-layer LSTM for temporal modeling.

Note that we use the proposed deep learning framework to estimate the locations of 14 key points/joints of the human body. These key points/joints include head, spine, left/right shoulder, left/right elbow, left/right wrist, left/right hip, left/right knee, and left/right ankle. We use a 3D skeleton composed of those key points/joints to represent the 3D human pose.

## 4.6 Loss Functions

In our system, we consider the problem of pose estimation as a regression problem that can directly regress the locations of joints of the human body. The training of the neural network can be considered as minimizing the average Euclidean distance error between predicted joints' locations and the ground truth. The loss function used for the training is composed of two parts including the position loss $L_P$ and the Huber loss $L_H$.

We first minimize the position loss $L_P$ that can be defined as the $L_2$ norm between the predicted joint position $\bar{p}_t^i$ and the ground truth $p_t^i$:

$$L_P = \frac{1}{T} \sum_{t=1}^{T} \frac{1}{N} \sum_{i=1}^{N} ||\bar{p}_t^i - p_t^i||_2, \qquad (10)$$

where $T$ means the input data sequence contains $T$ data samples and $N$ represents the number of joints. Next, we utilize the Huber loss $L_H$ that is a parameterized loss function for regression problems and less sensitive to outliers in data. The Huber loss can be defined as follows:

$$L_H = \frac{1}{T} \sum_{t=1}^{T} \frac{1}{N} \sum_{i=1}^{N} ||\bar{p}_t^i - p_t^i||_H, \qquad (11)$$

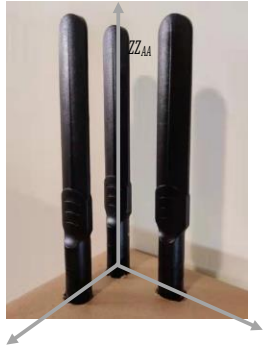
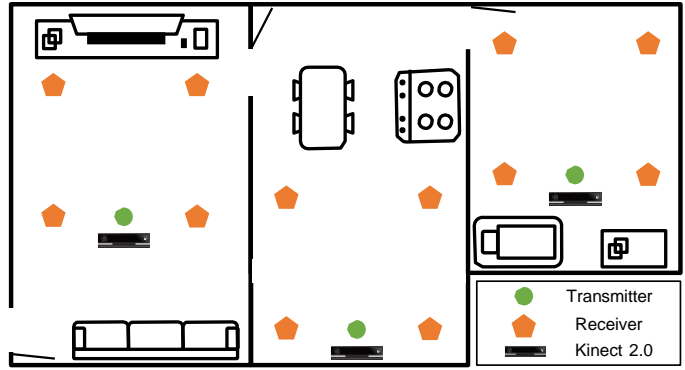

Fig. 9. The L-shaped antenna array at each receiver.

Fig. 10. Illustration of experimental environments and setup.

where $||\cdot||_H$ is the Huber norm [13].

With these two loss functions, we can write the overall objective function as follow:

$$L = Q_P \cdot L_P + Q_H \cdot L_H, \quad (12)$$

where $Q_P$ and $Q_H$ are the hyperparameters to balance the losses. We optimize the above objective function through Adam [16] in our system.

### 4.7 CSI Preprocessing

Before we feed the CSI measurements to the 2D AoA extraction component, we need to perform CSI preprocessing to clean the CSI noises. This is because the hardware imperfection of the commodity WiFi device results in CSI phase distortions. In particular, the receiver has a random phase shift due to sampling time offset (STO) and packet detection delay (PDD) across packets. We can apply a linear fit method proposed in [17] to remove the additional phase shift. Suppose $\psi(v, s, r)$ is the unwrapped phase of the CSI at the $v^{th}$ subcarrier of a packet transmitted from the $s^{th}$ transmitting antenna and received at the $r^{th}$ receiving antenna, we can obtain the optimal linear fit as follow:

$$\hat{\tau} = \mathrm{argmin} \sum_{v,s=1}^{V,S} \sum_{r=1}^{R} (\psi(v, s, r) + 2\pi f_\delta (r-1)\alpha + \beta)^2, \quad (13)$$

where $\alpha$ is the common slope of the received phase responses for all antennas and $\beta$ is the offset. The $\hat{\tau}$ includes the time delay of each packet and we can remove it to obtain the calibrated CSI phase with $\hat{\psi}(v, s, r) = \psi(v, s, r) - 2\pi f_\delta (r-1)\hat{\tau}$.

## 5. PERFORMANCE EVALUATION

### 5.1 Experimental Setup

**Devices.** We conduct the experiments with five laptops (one transmitter and four receivers). Each laptop runs Ubuntu 14.04 and is equipped with Intel 5300 wireless NIC connected to three antennas. The transmitter is equipped with three linearly-spaced antennas, and each receiver has an L-shaped antenna array as shown in Figure 9. Linux 802.11 CSI tools [8] are used to extract CSI measurements from 30 subcarriers for each packet.

Table 1. The average error of each joint (unit:*cm*).

| Joints | Head | Spine | LShoulder | LElbow | LWrist | RShoulder | RElbow | RWrist | LHip | LKnee | LAnkle | RHip | RKnee | RAnkle | Overall |
|---|---|---|---|---|---|---|---|---|---|---|---|---|---|---|---|
| CSI-based | 6.6 | 6.9 | 9.4 | 11.6 | 13.6 | 10.5 | 11.4 | 12.8 | 8.2 | 9.3 | 10.1 | 9.1 | 10.9 | 11.1 | 10.1 |
| GoPose | 3.2 | 3.0 | 3.9 | 6.0 | 8.4 | 4.8 | 5.1 | 7.2 | 3.9 | 3.1 | 4.7 | 3.3 | 4.6 | 5.1 | 4.7 |

The frequency band of the WiFi channel is 5.32 GHz with 40 MHz bandwidth. The default packet rate of our system is set at 1000 packets per second. We utilize a Microsoft Kinect 2.0 [27] to record the ground truth of the 3D human pose, and the sampling rate of Kinect 2.0 is set at 10 Hz. We use network time protocol (NTP) to ensure the synchronization for all the receivers and the Kinect 2.0.

**Environments.** We evaluate our system in three real-world environments in including a living room (4*m* × 4*m*), a dining room (3.6*m* × 3.6*m*), and a bedroom (4*m* × 3.8*m*). As shown in Figure 10, each room includes different furniture such as TV, sofa, table, bed, stove, etc. Figure 10 also shows the detailed deployments of all the devices in each environment. Note that the location of the furniture in each environment may change during the long-term experiments. If not specified, the default distance between two adjacent receivers is 2.5m. When the user is performing activities, she/he is also walking around freely in the area formed by the receivers. We place a Microsoft Kinect 2.0 to capture the participants and use frontal view data of Kinect without occlusion as the ground truth. It means that the Kinect can always capture the person in the line of sight.

**Model Setting.** We implement the stacked six-layer CNNs in our system, we use 3D convolution operation for both the 2D AoA spectrum and the CSI measurements. The number of the convolutional filters for the CNN layers are 64, 128, 256, 128, 64, and 1, respectively. The dropout rate is set as 0.1. In the LSTM, the number of the hidden state is set as 256 and the dropout rate is 0.1. The hyperparameters $Q\_p$ and $Q\_h$ in the loss function are 0.63 and 0.37, respectively. The networks are implemented in Keras.

**Data Collection.** In our experiments, 10 volunteers (including 6 males and 4 females) of various heights, weights, and ages are recruited. To evaluate the system performance, each volunteer is asked to conduct both exergaming activities according to the video tutorials and various everyday activities at will without any specific instruction while she/he is walking around. The exergaming activities include muscle-strengthening activities, balance and stretching exercises, aerobic exercises, dancing. The everyday activities include lifting arms, waving hands, walking, jogging, using a smartphone, etc. The total time span of our data collection is one month. We collect 676,200 samples of 3D skeleton frames and 67,620,000 WiFi CSI packets correspondingly in 3 different environments. The data set includes both one-person and two-person data. In particular, we split the dataset with 10 people into two non-overlapping datasets: a training set with 8 people and a test set with the other 2 people. This ensures that the training and test sets are from different people. We performed the leave-one-person-out cross-validation on the training set, where each time (i.e., each fold) we utilize 7 people for training and 1 person for validation. Thus, we can optimize the model parameters based on the 8-fold cross-validation. At last, we show the overall system performance using the test set and the trained model.

**Evaluation Metric.** We use the joint localization error as the evaluation metric. It is defined as the Euclidean distance between the predicted joint location and the ground truth. Note that we evaluate the 14 key points/joints mentioned in Section 4.5.

## 5.2 Overall Performance

We first show the overall performance of our system, and also compare it with the results of using CSI measurements as input of the deep learning framework (i.e., CSI-based approach). Figure 11(a) illustrates the cumulative distribution function (CDF) of joint localization errors for both GoPose and the CSI-based approach. We observe

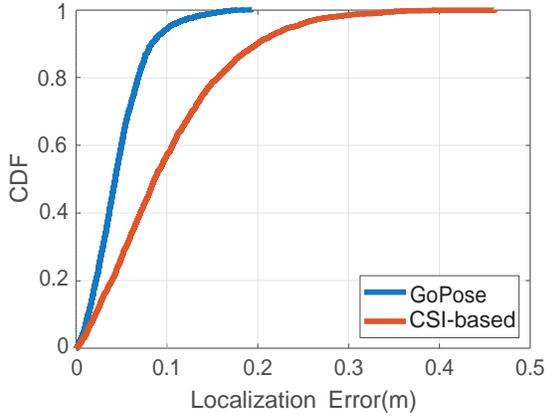
(a) Overall joint localization error.

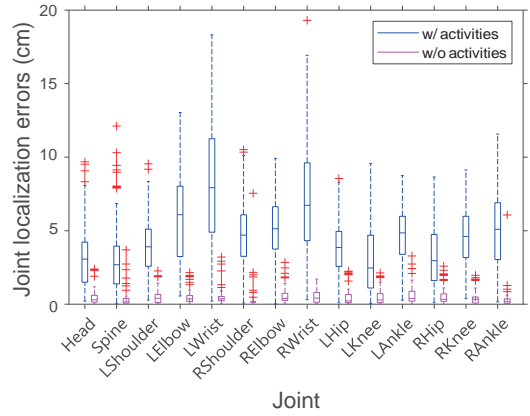
(b) The error distributions among different joints of both dynamic and static scenarios.

Fig. 11. Overall performance.

that GoPose can accurately track the 3D pose of human activities with an error of a few centimeters. Moreover, GoPhose significantly outperforms the CSI-based approach. In particular, the median joint estimation error of GoPose is 4.3cm, whereas it is 8.8cm for the CSI-based approach. Also, the 80 percentile joint estimation error is around 6.8cm for GoPose, while it is more than 15.6cm for the CSI-based approach. The reason that GoPose significantly outperforms the CSI-based approach is that the extracted 2D AoA spectrums can effectively capture the spatial information of the human body and are environment-independent, whereas CSI measurements provide no spatial features and are also affected by the background environments.

Table 1 reports the joint localization error for each joint for both GoPose and the CSI-based approach. We can find that the overall localization error of GoPose is only 4.7cm, whereas it is 10.1cm for the CSI-based approach. Additionally, the average joint localization error of GoPose ranges from 3.0cm to 8.4cm, while it ranges from 6.6cm to 13.6cm for the CSI-based approach. The results also show the significantly better performance of GoPhose. Also, the error distribution of each joint of GoPose is shown in Figure 11(b) with blue boxplot (i.e., w/ activities). All the medians are less than 8cm and most errors have a relatively small range. Moreover, we observe that the median and the range of the error of spine movements is much smaller than that of elbows, hands, and other joints. This is mainly because the torso has the largest reflection area among different joints, whereas the reflection of WiFi signal from arms is much weaker because of the smaller reflection areas. The accuracy of the joints with smaller reflection areas thus could be potentially improved by increasing the signal transmission power or directional antennas. We also show the error distribution of each joint when the user is not performing any activities. As we can see from the magenta boxplot (i.e., w/o activities) in Figure 11(b), all the medians are less than 0.4cm and all errors have a very small range. This is because it is relatively easier to track a stationary human body with few degrees of freedom.

To better visualize the performance of GoPose, we presented the constructed 3D human skeletons for different activities when the user is performing various activities. Figure 12 shows four examples of those constructed 3D skeletons. For each subfigure (e.g., Figure 12(a)), the first row shows the time series video frames recorded by the RGB camera for visual reference, and the second row shows the ground truth skeletons recorded by Kinect

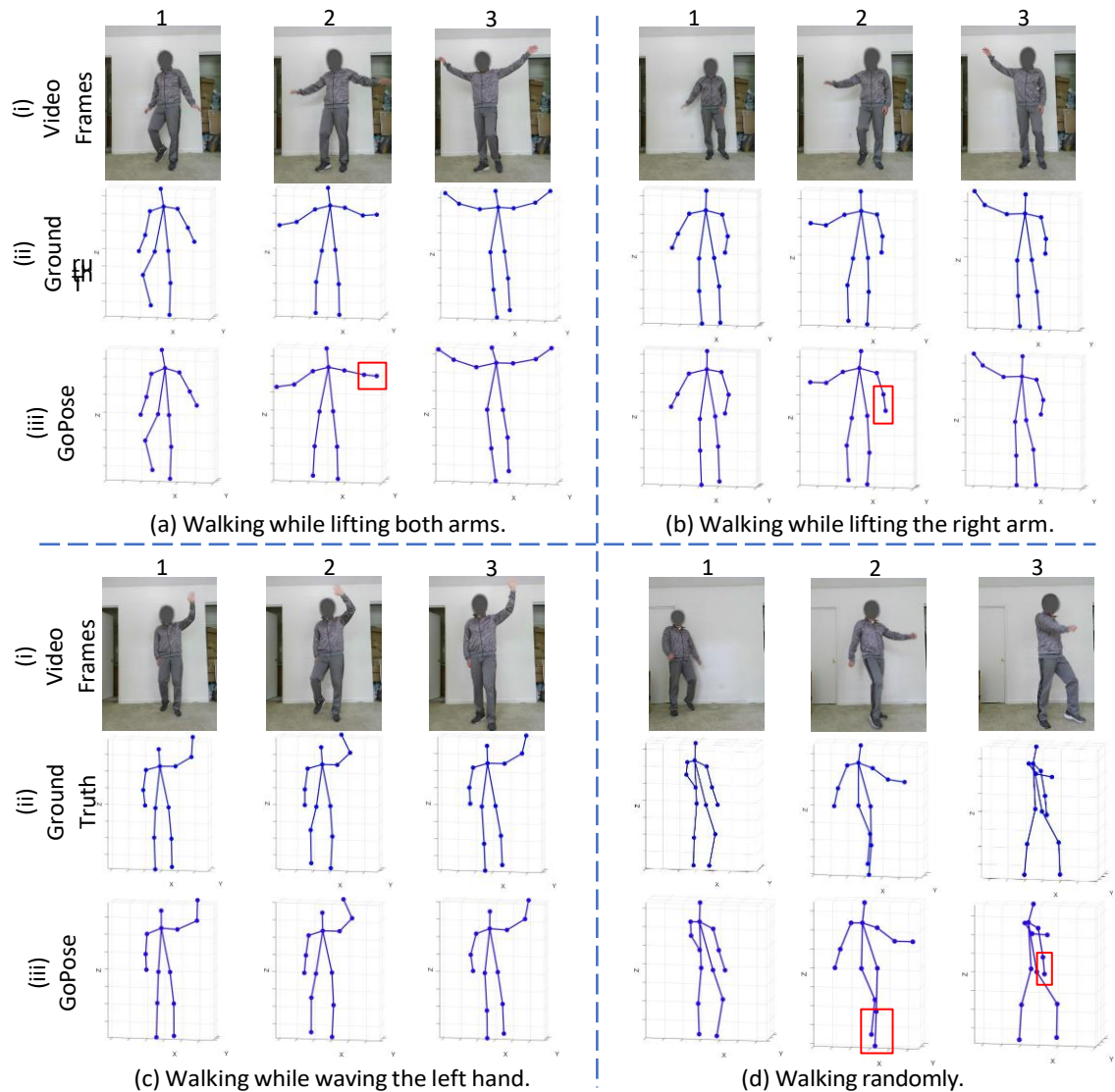

Fig. 12. Examples of the constructed skeletons.

2.0. The results of GoPose are shown in the third row. Figure 12(a), (b), and (c) show the user is performing different activities when he is walking. Figure 12(d) shows the user is randomly walking around. In addition, we use red solid rectangles to highlight the mispredicted and distorted body parts. For example, the $2^{nd}$ frame in Figure 12(a) has an inaccurate construction on the subject's left arm. We can see an incorrect left arm in the $2^{nd}$ frame in Figure 12(b). There are also a few slight deformations in Figure 12(d). Nevertheless, it is easy to observe that majority of the 3D poses estimated by GoPose are highly accurate. These results also demonstrate that the proposed system can accurately construct 3D moving human poses using WiFi signals.

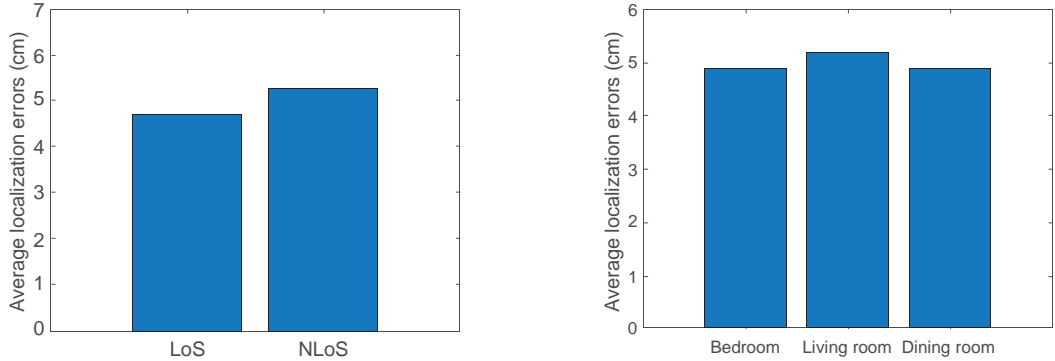

Fig. 13. System performance under both LoS and NLoS scenarios when the system is trained with LoS data.

Fig. 14. System performance in different environments when the system is trained in another environment.

### 5.3 Impact of Non-Line-of-Sight

To evaluate the effect of the NLoS scenario on the performance of our proposed system, we conduct experiments in two adjoining rooms separated by a wall. In particular, the WiFi transmitter and the person are in different rooms. Thus, the signals propagated to the human body and the signals reflected by the human body are obstructed by the wall. Note that we perform static environment removal in the evaluation. We utilize the trained model under the LoS scenario to test the performances of both LoS and NLoS as our system is environment-independent. The results of both LoS and NLoS scenarios are shown in Figure 13. We can observe that the average joint localization error under the LoS scenario is 4.7cm, whereas the error under the NLoS scenario is slightly higher (i.e., 5.3cm). Such a result demonstrates that our system can construct 3D moving human poses with high accuracy even under the NLoS scenario as the WiFi signals can penetrate obstacles. It also shows that our system can apply the deep learning model trained in LoS conditions to the NLoS scenarios without re-training. Therefore, GoPose is applicable to a wider range of applications, where the computer vision-based approaches could not work due to lack of line of sight or under poor lighting conditions.

### 5.4 Impact of Environmental Change

We have demonstrated that our system is able to achieve environment-independent where the performance stays the same when the location of the furniture was changed in the overall performance subsection. In this subsection, we further study the system performance under more challenging scenarios. In particular, we use the data collected in one environment (e.g., living room or dining room) to train our system, and then evaluate the performance when the system is operating in a different environment (e.g., bedroom). Note that we also perform static environment removal in this evaluation. Figure 14 shows the results for when operating the system in *Bedroom*, *Living room*, and *Dining room* respectively after the system was trained in a different environment. As shown in Figure 14, the average joint localization errors are 4.9cm, 5.2cm, and 4.8cm for *Bedroom*, *Living room*, and *Dining room*, respectively. Although testing environments are not seen during the training phase, the joint localization errors are still highly accurate. This is because our learning model relies on the 2D AoA spectrums of the signals only reflected by the human body, which is independent of the background environments. Our system thus is environmental-independent and can be trained in one environment and then operates in new environments without additional training.

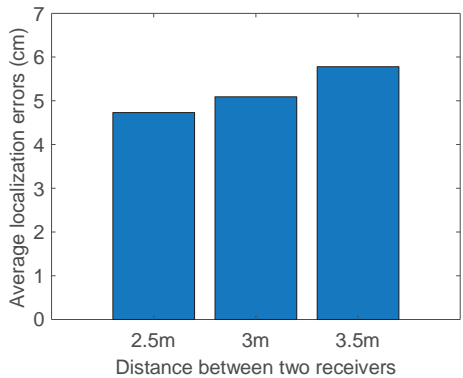
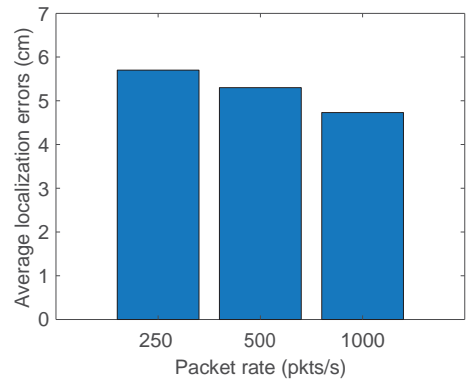

Fig. 15. System performance at different distances when the system is trained at the default distance.

Fig. 16. System performance under different packet rates when the system is trained with the default packet rate.

### 5.5 Impact of Different Distances

We next study the impact of the distance between the WiFi devices on the performance of our system. Here the distance is defined as the distance between two adjacent receivers and it may be adjusted by users according to the size of the room. We use the model trained at the default distance to evaluate the performance at different distances including 2.5m, 3m, and 3.5m. Figure 15 shows that the corresponding average errors are 4.7cm, 5.1cm, and 5.8cm, respectively. We can observe that system performance improves when the distance is reduced. This is because a shorter propagation distance leads to higher received signal strength. The results also show that the proposed system works well in the range of a typical room at various distance setups.

### 5.6 Impact of Packet Rate

For our evaluation, the default packet rate is set at 1000pkts/s. We further study the impact of different packet rates on the performance of our system. We set the packet rate at 250pkts/s, 500pkts/s and 1000pkts/s, respectively. As shown in Figure 16, the average errors for 250pkts/s, 500pkts/s and 1000pkts/s are 5.7cm, 5.3cm and 4.7cm, respectively. We can observe that a higher packet rate slightly improves system performance. The reason is twofold: first, a high packet rate can easily recover the high-speed movements of the body part; second, we can combine more packets to better represent the full-body movements of a user when using a higher packet rate. Still, GoPose works relatively well even under low packet rates (e.g., 250pkts/s).

### 5.7 Impact of Subject Diversity

We next evaluate the subject diversity on the performance of our system since the system could be used by a diverse set of users. We have a training set with 8 people and we conduct the evaluation using leave-one-person-out cross-validation on these people. Thus, each time we utilize 7 people for training and 1 person for validation. The results are shown in Figure 17. For example, the result of subject 1 means that the model was trained with the data from subject 2 to subject 8, and was validated with the data of subject 1. We can observe that all joint localization errors are less than 4.9cm, and there are no significant differences among these subjects. The average error of the leave-one-person-out cross-validation is only 4.6cm. This demonstrates our system can achieve high accuracy when used by different users and the model is not overfitting.

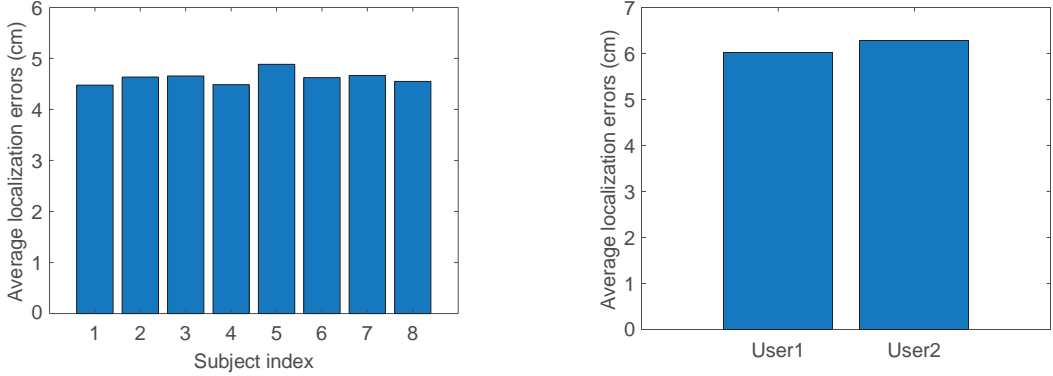

Fig. 17. System performance for different subjects based on leave-one-person-out cross-validation.

Fig. 18. System performance under the multi-user scenario.

### 5.8 Impact of Multiple Users

In this subsection, we focus on the impact of multiple users by asking two users to walk and perform activities simultaneously. We first detect the number of users in the environment by taking the 2D AoA as input to a simple CNN classifier. This is because the multi-user scenarios tend to have more multiple human bodies reflected signals that result in multiple body shapes in the 2D AoA spectrum. Thus, it is easy to recognize the number of users if there is no overlapping human body. In the meanwhile, different model parameters are specifically trained with different numbers of users. After we detect the number of users, we utilize the model with the corresponding parameters for pose estimation. In our evaluation, we test the multi-user scenario with two people in one room. In particular, if the system detects two people, we choose the two-user model (i.e., the model with two-user parameters) to estimate the poses. Figure 18 shows the average errors for each of the two users, i.e., *user1* and *user2*. We observe that the overall error is 6.0cm and 6.3cm for *user1* and *user2*, respectively. The results are slightly worse than that of the overall performance of the system. This is because multiple users can lead to complex signal reflections, which will result in slightly larger errors. We believe that with more WiFi devices, the complex signal reflections could be resolved better. Increasing the number of WiFi devices thus can improve the system performance under multiple user scenarios. Still, our current system works well when there are two users performing activities simultaneously in the home environment.

### 5.9 Ablation Study

We perform ablation studies to evaluate the contribution of each component of our system by either removing or changing one component while others remain the same.

Figure 19(a) shows the impact of the number of receivers on the system performance. We can observe that the errors of the one-receiver scenario, two-receiver scenario, three-receiver scenario, four-receiver scenario are 8cm, 6cm, 5.2cm, and 4.7cm, respectively. We observe that the error of the one-receiver scenario is much larger than the others and using more receivers can improve performance slightly. Thus, our system could achieve comparable performance with only 2 receivers in a typical smart home environment, in where multiple WiFi transceivers may be deployed.

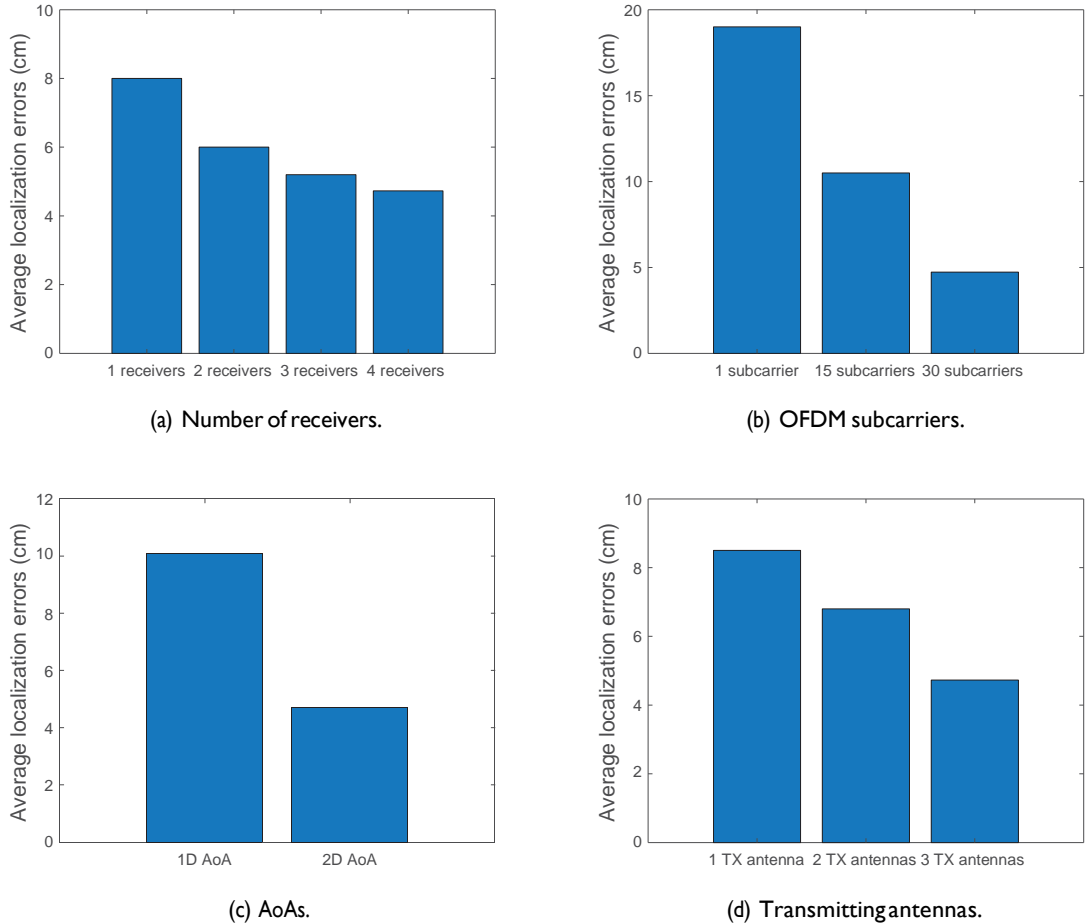

Fig. 19. Ablation studies.

Figure 19(b) represents how the system performance varies with the number of OFDM subcarriers. Compared to using only one subcarrier (the error is about 19cm), system performance is significantly increased (the error is 4.7cm) when we leverage all 30 OFDM subcarriers. It is because a number of subcarriers provide more frequency diversities and thus can improve the 2D AoA resolution.

As shown in Figure 19(c), using 2D AoA achieves a much better performance which has an error of 4.7cm, whereas the 1D AoA achieves an error of 10.1cm. The reason is that the 2D AoA provides more dimensional information than 1D AoA does.

Figure 19(d) illustrates the effectiveness of multiple transmitting antennas. The errors of using 1, 2, and 3 transmitting antennas are 8.5cm, 6.8cm, and 4.7cm, respectively. This is because multiple transmitting antennas can increase the spatial diversity at the transmitter. According to these ablation studies, we can find that all these components could contribute to the system performance, in which multiple OFDM subcarriers and 2D AoA play important roles.

## 6 DISCUSSION

Although GoPose can achieve on-the-go 3D pose estimation for unseen activities with high accuracy in various scenarios, it still has some limitations.

**Pose Estimation for Multiple Users.** Currently, our system only tested one or two people in a typical room environment. We acknowledge that the pose estimation for a large number of users (i.e., more than 3 people) in one room is still a challenge. It is because the human bodies could overlap in such a crowded environment. Such an issue also occurs in the computer vision community. In addition, a large number of users would lead to more complicated signal reflections.

**User Study.** There are a limited number of subjects participating in our experiments as it is challenging to find a large number of volunteers to conduct the experiments due to the impact of the COVID-19 pandemic. A more comprehensive user study over a larger number of users could better evaluate our system. We would like to include more users in our future work.

**Long-Range Pose Estimation.** Although we improve the performance of 2D AoA with the spatial diversity at transmitting antennas and the frequency diversity of OFDM subcarriers, the sensing range of our system could be still limited. As the distance becomes greater, it becomes more difficult for the system to distinguish between two objects with a fixed spatial resolution.

## 7 CONCLUSION

This paper presents GoPose, a 3D skeleton-based human pose estimation system that offers on-the-go pose tracking for unseen activities in a home environment. It analyzes the WiFi signal bounced off the human body for 3D pose estimation and can reuse the WiFi devices that already exist at home for mass adoption. In the GoPose system, the 2D AoA spectrum of the signals reflected from the human body is leveraged to locate different parts of the human body as well as to enable environment-independent sensing, while deep learning is incorporated to model the complex relationship between the 2D AoA spectrums and the 3D skeletons of the human body for 3D pose tracking. We evaluate GoPose in different home environments with various activities performed by multiple users. The evaluation shows that GoPose is environment-independent and is highly accurate in constructing 3D human poses for mobile users. Results also show that GoPose achieves around 4.7cm of accuracy under various scenarios including tracking unseen activities and under NLoS scenarios.

## REFERENCES

[1] Karan Ahuja, Sven Mayer, Mayank Goel, and Chris Harrison. 2021. Pose-on-the-Go: Approximating User Pose with Smartphone Sensor Fusion and Inverse Kinematics. In *Proceedings of the 2021 CHI Conference on Human Factors in Computing Systems*. 1–12.

[2] Teo Babic, Florian Perteneder, Harald Reiterer, and Michael Haller. 2020. Simo: Interactions with distant displays by smartphones with simultaneous face and world tracking. In *Extended Abstracts of the 2020 CHI Conference on Human Factors in Computing Systems*. 1–12.

[3] Alan Bränzel, Christian Holz, Daniel Hoffmann, Dominik Schmidt, Marius Knaust, Patrick Lühne, René Meusel, Stephan Richter, and Patrick Baudisch. 2013. GravitySpace: tracking users and their poses in a smart room using a pressure-sensing floor. In *Proceedings of the SIGCHI Conference on Human Factors in Computing Systems*. 725–734.

[4] Zhe Cao, Tomas Simon, Shih-En Wei, and Yaser Sheikh. 2017. Realtime multi-person 2d pose estimation using part affinity fields. In *Proceedings of the IEEE conference on computer vision and pattern recognition*. 7291–7299.

[5] Ke-Yu Chen, Shwetak N Patel, and Sean Keller. 2016. Finexus: Tracking precise motions of multiple fingertips using magnetic sensing. In *Proceedings of the 2016 CHI Conference on Human Factors in Computing Systems*. 1504–1514.

[6] Mahmoud El-Gohary and James McNames. 2012. Shoulder and elbow joint angle tracking with inertial sensors. *IEEE Transactions on Biomedical Engineering* 59, 9 (2012), 2635–2641.

[7] Halfbrick. 2021. Fruit Ninja VR. https://www.halfbrick.com/games/fruit-ninja-vr

[8] Daniel Halperin, Wenjun Hu, Anmol Sheth, and David Wetherall. 2011. Tool release: Gathering 802.11n traces with channel state information. *ACM SIGCOMM Computer Communication Review* 41, 1 (2011), 53–53.

[9] Ferid Harabi, Ali Gharsallah, and Sylvie Marcos. 2009. Three-dimensional antennas array for the estimation of direction of arrival. *IET microwaves, antennas & propagation* 3, 5 (2009), 843–849.

[10] Kaiming He, Georgia Gkioxari, Piotr Dollár, and Ross Girshick. 2017. Mask r-cnn. In *Proceedings of the IEEE international conference on computer vision*. 2961–2969.

[11] Dun-Yu Hsiao, Min Sun, Christy Ballweber, Seth Cooper, and Zoran Popović. 2016. Proactive sensing for improving hand pose estimation. In *Proceedings of the 2016 CHI Conference on Human Factors in Computing Systems*. 2348–2352.

[12] Stephen S Intille, Ling Bao, Emmanuel Munguia Tapia, and John Rondoni. 2004. Acquiring in situ training data for context-aware ubiquitous computing applications. In *Proceedings of the SIGCHI conference on Human factors in computing systems*. 1–8.

[13] Wenjun Jiang, Hongfei Xue, Chenglin Miao, Shiyang Wang, Sen Lin, Chong Tian, Srinivasan Murali, Haochen Hu, Zhi Sun, and Lu Su. 2020. Towards 3D human pose construction using wifi. In *Proceedings of the 26th Annual International Conference on Mobile Computing and Networking*. 1–14.

[14] Angjoo Kanazawa, Michael J Black, David W Jacobs, and Jitendra Malik. 2018. End-to-end recovery of human shape and pose. In *Proceedings of the IEEE Conference on Computer Vision and Pattern Recognition*. 7122–7131.

[15] Cagdas Karatas, Luyang Liu, Hongyu Li, Jian Liu, Yan Wang, Sheng Tan, Jie Yang, Yingying Chen, Marco Gruteser, and Richard Martin. 2016. Leveraging wearables for steering and driver tracking. In *IEEE INFOCOM 2016-The 35th Annual IEEE International Conference on Computer Communications*. IEEE, 1–9.

[16] Diederik P Kingma and Jimmy Ba. 2014. Adam: A method for stochastic optimization. *arXiv preprint arXiv:1412.6980* (2014).

[17] Manikanta Kotaru, Kiran Joshi, Dinesh Bharadia, and Sachin Katti. 2015. Spotfi: Decimeter level localization using wifi. In *Proceedings of the 2015 ACM Conference on Special Interest Group on Data Communication*. 269–282.

[18] Jooshik Lee, Iickho Song, Hyoungmoon Kwon, and Sung Ro Lee. 2003. Low-complexity estimation of 2D DOA for coherently distributed sources. *Signal processing* 83, 8 (2003), 1789–1802.

[19] Hanchuan Li, Peijin Zhang, Samer Al Moubayed, Shwetak N Patel, and Alanson P Sample. 2016. Id-match: A hybrid computer vision and rfid system for recognizing individuals in groups. In *Proceedings of the 2016 CHI Conference on Human Factors in Computing Systems*. 4933–4944.

[20] Jianfeng Li, Penghui Ma, Xiaofei Zhang, and Gaofeng Zhao. 2020. Improved DFT algorithm for 2D DOA estimation based on 1D nested array motion. *IEEE Communications Letters* 24, 9 (2020), 1953–1956.

[21] Tianxing Li, Chuankai An, Zhao Tian, Andrew T Campbell, and Xia Zhou. 2015. Human sensing using visible light communication. In *Proceedings of the 21st Annual International Conference on Mobile Computing and Networking*. 331–344.

[22] Xiang Li, Shengjie Li, Daqing Zhang, Jie Xiong, Yasha Wang, and Hong Mei. 2016. Dynamic-music: accurate device-free indoor localization. In *Proceedings of the 2016 ACM International Joint Conference on Pervasive and Ubiquitous Computing*. 196–207.

[23] Hongbo Liu, Yu Gan, Jie Yang, Simon Sidhom, Yan Wang, Yingying Chen, and Fan Ye. 2012. Push the limit of WiFi based localization for smartphones. In *Proceedings of the 18th annual international conference on Mobile computing and networking*. 305–316.

[24] Jian Liu, Yingying Chen, Yan Wang, Xu Chen, Jerry Cheng, and Jie Yang. 2018. Monitoring vital signs and postures during sleep using WiFi signals. *IEEE Internet of Things Journal* 5, 3 (2018), 2071–2084.

[25] Jian Liu, Yan Wang, Yingying Chen, Jie Yang, Xu Chen, and Jerry Cheng. 2015. Tracking vital signs during sleep leveraging off-the-shelf wifi. In *Proceedings of the 16th ACM International Symposium on Mobile Ad Hoc Networking and Computing*. 267–276.

[26] Dushyant Mehta, Srinath Sridhar, Oleksandr Sotnychenko, Helge Rhodin, Mohammad Shafiei, Hans-Peter Seidel, Weipeng Xu, Dan Casas, and Christian Theobalt. 2017. Vnect: Real-time 3d human pose estimation with a single rgb camera. *ACM Transactions on Graphics (TOG)* 36, 4 (2017), 1–14.

[27] Microsoft. 2021. Kinect 2 for Windows. https://developer.microsoft.com/en-us/windows/kinect/

[28] Dan Morris, T Scott Saponas, Andrew Guillory, and Ilya Kelner. 2014. RecoFit: using a wearable sensor to find, recognize, and count repetitive exercises. In *Proceedings of the SIGCHI Conference on Human Factors in Computing Systems*. 3225–3234.

[29] George Papandreou, Tyler Zhu, Nori Kanazawa, Alexander Toshev, Jonathan Tompson, Chris Bregler, and Kevin Murphy. 2017. Towards accurate multi-person pose estimation in the wild. In *Proceedings of the IEEE Conference on Computer Vision and Pattern Recognition*. 4903–4911.

[30] Dario Pavllo, Christoph Feichtenhofer, David Grangier, and Michael Auli. 2019. 3d human pose estimation in video with temporal convolutions and semi-supervised training. In *Proceedings of the IEEE/CVF Conference on Computer Vision and Pattern Recognition*. 7753–7762.

[31] Qifan Pu, Sidhant Gupta, Shyamnath Gollakota, and Shwetak Patel. 2013. Whole-home gesture recognition using wireless signals. In *Proceedings of the 19th annual international conference on Mobile computing & networking*. 27–38.

[32] Kun Qian, Chenshu Wu, Zimu Zhou, Yue Zheng, Zheng Yang, and Yunhao Liu. 2017. Inferring motion direction using commodity wi-fi for interactive exergames. In *Proceedings of the 2017 CHI Conference on Human Factors in Computing Systems*. 1961–1972.

[33] Yanzhi Ren, Yingying Chen, Mooi Choo Chuah, and Jie Yang. 2013. Smartphone based user verification leveraging gait recognition for mobile healthcare systems. In *2013 IEEE international conference on sensing, communications and networking (SECON)*. IEEE, 149–157.

[34] Yanzhi Ren, Yingying Chen, Mooi Choo Chuah, and Jie Yang. 2014. User verification leveraging gait recognition for smartphone enabled mobile healthcare systems. *IEEE Transactions on Mobile Computing* 14, 9 (2014), 1961–1974.

[35] Yili Ren, Sheng Tan, Linghan Zhang, Zi Wang, Zhi Wang, and Jie Yang. 2020. Liquid Level Sensing Using Commodity WiFi in a Smart Home Environment. *Proceedings of the ACM on Interactive, Mobile, Wearable and Ubiquitous Technologies* 4, 1 (2020), 1–30.
[36] Yili Ren, Zi Wang, Sheng Tan, Yingying Chen, and Jie Yang. 2021. Tracking free-form activity using wifi signals. In *Proceedings of the 27th Annual International Conference on Mobile Computing and Networking*. 816–818.
[37] Yili Ren, Zi Wang, Sheng Tan, Yingying Chen, and Jie Yang. 2021. Winect: 3D Human Pose Tracking for Free-form Activity Using Commodity WiFi. *Proceedings of the ACM on Interactive, Mobile, Wearable and Ubiquitous Technologies* 5, 4 (2021), 1–29.
[38] Ralph Schmidt. 1986. Multiple emitter location and signal parameter estimation. *IEEE transactions on antennas and propagation* 34, 3 (1986), 276–280.
[39] Sheng Shen, He Wang, and Romit Roy Choudhury. 2016. I am a smartwatch and i can track my user's arm. In *Proceedings of the 14th annual international conference on Mobile systems, applications, and services*. 85–96.
[40] Leonid Sigal, Alexandru O Balan, and Michael J Black. 2010. Humaneva: Synchronized video and motion capture dataset and baseline algorithm for evaluation of articulated human motion. *International journal of computer vision* 87, 1-2 (2010), 4.
[41] Sheng Tan, Yili Ren, Jie Yang, and Yingying Chen. 2022. Commodity WiFi Sensing in 10 Years: Status, Challenges, and Opportunities. *IEEE Internet of Things Journal* (2022).
[42] Sheng Tan and Jie Yang. 2016. WiFinger: Leveraging commodity WiFi for fine-grained finger gesture recognition. In *Proceedings of the 17th ACM international symposium on mobile ad hoc networking and computing*. 201–210.
[43] Sheng Tan, Jie Yang, and Yingying Chen. 2020. Enabling fine-grained finger gesture recognition on commodity wifi devices. *IEEE Transactions on Mobile Computing* (2020).
[44] Sheng Tan, Linghan Zhang, Zi Wang, and Jie Yang. 2019. MultiTrack: Multi-user tracking and activity recognition using commodity WiFi. In *Proceedings of the 2019 CHI Conference on Human Factors in Computing Systems*. 1–12.
[45] Sheng Tan, Linghan Zhang, and Jie Yang. 2018. Sensing fruit ripeness using wireless signals. In *2018 27th International Conference on Computer Communication and Networks (ICCCN)*. IEEE, 1–9.
[46] Maksym Tatariants. 2021. Human Pose Estimation Technology 2021 Guide. https://mobidev.biz/blog/human-pose-estimation-ai-personal-fitness-coach
[47] Jochen Tautges, Arno Zinke, Björn Krüger, Jan Baumann, Andreas Weber, Thomas Helten, Meinard Müller, Hans-Peter Seidel, and Bernd Eberhardt. 2011. Motion reconstruction using sparse accelerometer data. *ACM Transactions on Graphics (ToG)* 30, 3 (2011), 1–12.
[48] Ultraleap. 2021. Leap Motion Controller. https://www.ultraleap.com/product/leap-motion-controller/
[49] Elise Klæbo Vonstad, Xiaomeng Su, Beatrix Vereijken, Kerstin Bach, and Jan Harald Nilsen. 2020. Comparison of a Deep Learning-Based Pose Estimation System to Marker-Based and Kinect Systems in Exergaming for Balance Training. *Sensors* 20, 23 (2020), 6940.
[50] Chuyu Wang, Jian Liu, Yingying Chen, Lei Xie, Hong Bo Liu, and Sanclu Lu. 2018. RF-kinect: A wearable RFID-based approach towards 3D body movement tracking. *Proceedings of the ACM on Interactive, Mobile, Wearable and Ubiquitous Technologies* 2, 1 (2018), 1–28.
[51] Fei Wang, Sanping Zhou, Stanislav Panev, Jinsong Han, and Dong Huang. 2019. Person-in-WiFi: Fine-grained person perception using WiFi. In *Proceedings of the IEEE/CVF International Conference on Computer Vision*. 5452–5461.
[52] Xianpeng Wang, Mengxing Huang, and Liangtian Wan. 2021. Joint 2D-DOD and 2D-DOA estimation for coprime EMVS–MIMO radar. *Circuits, Systems, and Signal Processing* 40, 6 (2021), 2950–2966.
[53] Yan Wang, Jian Liu, Yingying Chen, Marco Gruteser, Jie Yang, and Hongbo Liu. 2014. E-eyes: device-free location-oriented activity identification using fine-grained wifi signatures. In *Proceedings of the 20th annual international conference on Mobile computing and networking*. 617–628.
[54] Yinsheng Wei and Xiaojiang Guo. 2014. Pair-matching method by signal covariance matrices for 2D-DOA estimation. *IEEE Antennas and Wireless Propagation Letters* 13 (2014), 1199–1202.
[55] Erwin Wu, Ye Yuan, Hui-Shyong Yeo, Aaron Quigley, Hideki Koike, and Kris M Kitani. 2020. Back-Hand-Pose: 3D Hand Pose Estimation for a Wrist-worn Camera via Dorsum Deformation Network. In *Proceedings of the 33rd Annual ACM Symposium on User Interface Software and Technology*. 1147–1160.
[56] Jie Yang and Yingying Chen. 2008. A theoretical analysis of wireless localization using RF-based fingerprint matching. In *2008 IEEE International Symposium on Parallel and Distributed Processing*. IEEE, 1–6.
[57] Jie Yang and Yingying Chen. 2009. Indoor localization using improved rss-based lateration methods. In *GLOBECOM 2009 IEEE Global Telecommunications Conference*. IEEE, 1–6.
[58] Youwei Zeng, Dan Wu, Jie Xiong, Jinyi Liu, Zhaopeng Liu, and Daqing Zhang. 2020. MultiSense: Enabling multi-person respiration sensing with commodity wifi. *Proceedings of the ACM on Interactive, Mobile, Wearable and Ubiquitous Technologies* 4, 3 (2020), 1–29.
[59] Feng Zhang, Chenshu Wu, Beibei Wang, and KJ Ray Liu. 2020. mmEye: Super-resolution millimeter wave imaging. *IEEE Internet of Things Journal* 8, 8 (2020), 6995–7008.
[60] Yang Zhang, Chouchang Yang, Scott E Hudson, Chris Harrison, and Alanson Sample. 2018. Wall++ room-scale interactive and context-aware sensing. In *Proceedings of the 2018 CHI Conference on Human Factors in Computing Systems*. 1–15.
[61] Zhengyou Zhang. 2012. Microsoft kinect sensor and its effect. *IEEE multimedia* 19, 2 (2012), 4–10.

[62] Mingmin Zhao, Tianhong Li, Mohammad Abu Alsheikh, Yonglong Tian, Hang Zhao, Antonio Torralba, and Dina Katabi. 2018. Through-wall human pose estimation using radio signals. In *Proceedings of the IEEE Conference on Computer Vision and Pattern Recognition*. 7356–7365.

[63] Mingmin Zhao, Yonglong Tian, Hang Zhao, Mohammad Abu Alsheikh, Tianhong Li, Rumen Hristov, Zachary Kabelac, Dina Katabi, and Antonio Torralba. 2018. RF-based 3D skeletons. In *Proceedings of the 2018 Conference of the ACM Special Interest Group on Data Communication*. 267–281.

[64] Anastasiya Zharovskikh. 2020. Pose Estimation to Empower Your Business. https://indatalabs.com/blog/pose-estimation

[65] Xiuyuan Zheng, Hongbo Liu, Jie Yang, Yingying Chen, Richard P Martin, and Xiaoyan Li. 2013. A study of localization accuracy using multiple frequencies and powers. *IEEE Transactions on Parallel and Distributed Systems* 25, 8 (2013), 1955–1965.